# SLM4Offer: Personalized Marketing Offer Generation Using Contrastive Learning Based Fine-Tuning

Vedasamhitha Challapalli, Konduru Venkat Sai, Piyush Pratap Singh, Rupesh Prasad, Arvind Maurya, and Atul Singh

SPARC Research, HCLSoftware, Bengaluru

**Abstract.** Personalized marketing has emerged as a pivotal strategy for enhancing customer engagement and driving business growth. Academic and industry efforts have predominantly focused on recommendation systems and personalized advertisements. Nonetheless, this facet of personalization holds significant potential for increasing conversion rates and improving customer satisfaction. Prior studies suggest that well-executed personalization strategies can boost revenue by up to 40%, underscoring the strategic importance of developing intelligent, data-driven approaches for offer generation.

This work introduces SLM4Offer, a generative AI model for personalized offer generation, developed by fine-tuning a pre-trained encoder-decoder language model—specifically Google's Text-to-Text Transfer Transformer (T5-Small 60M) —using a contrastive learning approach. SLM4Offer employs InfoNCE (Information Noise-Contrastive Estimation) loss to align customer personas with relevant offers in a shared embedding space.

A key innovation in SLM4Offer lies in the adaptive learning behavior introduced by contrastive loss, which reshapes the latent space during training and enhances the model's generalizability The model is fine-tuned and evaluated on a synthetic dataset designed to simulate customer behavior and offer acceptance patterns. Experimental results demonstrate a 17% improvement in offer acceptance rate over a supervised fine-tuning baseline, highlighting the effectiveness of contrastive objectives in advancing personalized marketing.

## 1 Introduction

Personalized marketing has been shown to significantly impact business performance. Arora et al. (2021) [1] report that firms can achieve up to 40% increase in revenue through the use of personalized marketing strategies. Previous research has also highlighted various positive consumer responses to personalization(de Groot, 2022; Tomczyk, Buhalis, Fan, & Williams, 2022)[7]. According to Kotler and Keller (Marketing Management, 15th Edition), 'a marketing offer is defined as a combination of products, services, information, or experiences presented to a market to fulfill a specific need or want'. This paper presents SLM4Offer, a Gen AI model, built by applying contrastive learning based fine-tuning on Google's T5, to create personalized marketing offers for a given customer profile.

This approach builds on the foundational work of Zhang et al., 2022 [8], where contrastive learning was used to align word embeddings with their definitions at a fine semantic level. In contrast to their work, which considers the definitions of all other words as negatives, this paper applies contrastive learning to hard negatives— which are the rejected offers by a particular customer. To the best of the author's knowledge this is the first time that contrastive fine-tuning of a Generative AI model is used to generate personalized marketing offers.
This paper compares the performance of contrastive learning-based fine-tuning with a baseline approach that employs supervised fine-tuning on the same Google T5 model. The authors report with 94% certainty that LLM's judgments align with human evaluations using the standard chi-square test.The simulation results indicate that contrastive loss-based fine-tuning achieves approximately 17% improvement in the acceptance rate—defined as the ratio of offers accepted to the total offers proposed - compared to the baseline approach.

The remainder of this paper is organized as follows. Section 2 reviews related work on the application of Large Language Models (LLMs) and contrastive learning. Section 3 presents the necessary background to support a deeper understanding of our approach. Section 4 details the fine-tuning process used to create SLM4Offer. Section 5 outlines the experimental methodology to evaluate SLM4Offer, followed by the results and analysis in Section 6, followed by the section Conclusion which concludes the paper and discusses potential directions for future work.

## 2  Related Work

The existing approaches for marketing offer generation can be broadly divided into three categories. The first category consists of retrieval-based recommenders, which select offers from a predefined set of offers using similarity to historical acceptance data (Ni et al., 2017 [4]). The second category consists of classification-based systems, which assign likelihood scores to a fixed pool of offers based on customer profiles. The last category of offers consists of generation-based approaches, which aim to synthesize new and personalized offers from scratch. The work presented in this paper belongs to the last category and introduces a novel contrastive training mechanism within a generative model. The approach presented in this paper builds on the definition generation task, which was introduced by Noraset et al. (2017)[5]. This work was extended by Huang et al. (2021)[2] and Kong et al. (2022)[3], who employed pre-trained encoder–decoder models, achieving improved fluency and contextual relevance. The work presented in this paper builds upon InfoNCE which is a contrastive loss introduced by Aaron Van Den Oord et al. [6].

In summary, while prior work in offer generation either reuses existing offers or generates new ones through generic language models, the paper introduces a novel contrastive training mechanism integrated within a generative model, specifically tailored to marketing personalization.

## 3 Background

This section presents an overview of Google's Text-To-Text Transfer Transformer (T5) small language model (SLM) used in this paper and provides an outline of the contrastive learning Information Noise-Contrastive Estimation (InfoNCE) loss.

### 3.1 Text-to-Text Transfer Transformer(T5)

Large language models (LLMs) are transformer based models trained on immense amounts of data making them capable of understanding and generating natural language content for a wide range of tasks. A Small Language Model (SLM) is an LLM with lesser number of parameters compared to larger LLMs. There is no agreed definition of the size of a Small Language Model (SLM) in literature.

Developed by Google Research, T5 (Text-to-Text Transfer Transformer) is an encoder-decoder model trained on the large-scale text corpus called C4 (Colossal Clean Crawled Corpus). The model is available in multiple sizes, ranging from 60M to 11B parameters, and supports a maximum input length of 512 to 1024 tokens. The authors have chosen a T5 small (60M) encoder-decoder model due to its ability to effectively capture input-output relationships, especially when the input (customer context) and output (personalized offers) differ structurally thereby making it well-suited for generating personalized responses conditioned on diverse customer profiles with their past history of accepted and rejected offers. SLM4Offer is scalable to larger T5 variants and other LLMs also. The smaller model was intentionally chosen to test the algorithm in a constrained setup with minimal domain knowledge.

### 3.2 Contrastive Learning and InfoNCE loss

In this paper the authors use **contrastive learning**-based Information Noise-Contrastive Estimation (InfoNCE) loss function introduced by Aaron Van Den Oord et al [6] to effectively capture the distinctions between accepted and rejected offers in relation to customer personas. In the approach presented in this paper, the authors define a **positive pair** as the combination of a customer persona and the accepted offers by the customer, and a **negative pair** as the customer persona and the rejected offers.

InfoNCE loss is preferred over other contrastive losses such as triplet loss because it is more efficient especially in setups like in the paper, where it is required to embed customers and offers in a shared space and need a strong signal to bring positives closer while pushing all negatives away. Whereas, Triplet loss lacks this deep probabilistic grounding, making it harder to optimize.

The InfoNCE loss is formulated as:

$$P(a_j \mid z_i) = \frac{\exp\left(\frac{sim(z_i, a_j)}{\tau}\right)}{\sum_{k=1}^{N} \exp\left(\frac{sim(z_i, r_k)}{\tau}\right)}$$

$$\mathcal{L}_{InfoNCE} = -E_X \left[\log \left(P(a_j \mid z_i)\right)\right]$$

where:

- $sim(u, v)$ represents the cosine similarity between two vectors $u$ and $v$,
- $T$ is a temperature hyperparameter controlling the sharpness of the distribution, $N$ is the number of negative samples in the batch.
- $z_i$ is the representation of a data point, $a_j$ is the representation of positive sample, $r_i$ is the representation of negative sample

## 4   SLM4Offer: Marketing Offer Generation

This section begins by explaining how SLM4Offer generates personalized marketing offers. It then introduces the dual-objective loss function used to fine-tune the T5 encoder-decoder architecture on which SLM4Offer is built. The model takes a prompt and a customer persona as input to produce a personalized marketing offer. Figure 1 presents sample persona and offers generated by SLM4Offer.

SLM4Offer uses a dual objective loss function which is given by:

$$L_{Final} = \lambda \cdot L_C + (1 - \lambda) \cdot L_G$$

Here, $L_C$ denotes the contrastive loss as described in Section 3 and $L_G$ represents the cross-entropy-based offer generation loss. The value of $\lambda$ depends on the model, the specific use case, and the desired extent of unsupervised training in the embedding space. A value of 0.5 has been used as the $\lambda$ for the current scenario. This dual loss function ensures that while the decoder learns to generate

**Customer Persona**

```
{"Name": "P9654",
 "Age": 30,
 "Gender": "Female",
 "Monthly Income": "$111,969",
 "Spending Pattern": "Budget-conscious",
 "Preferred Payment Method": "Buy Now Pay Later",
 "Interests": ["Finance", "Fitness", "Gaming"],
 "Financial Goals": ["Wealth Growth", "Retirement
  Planning"]}
```

```
{"Name": "P9999",
 "Age": 35,
 "Gender": "Male",
 "Monthly Income": "$82,450",
 "Spending Pattern": "High-spender",
 "Preferred Payment Method": "Credit Card",
 "Interests": ["Fitness", "Shopping", "Technology",
  "Travel"],
 "Financial Goals": ["Savings", "Wealth Growth"]}
```

**Generated Offers by SLM4Offer**

Investment Planning Services for Wealth Growth and Retirement Planning.

High-End Fitness Equipment with 5% Cashback on Online Shopping.

**Fig. 1.** Example of customer persona and offers generated by SLM4Offer

offers in a supervised manner using the cross-entropy loss, the embedding space is simultaneously trained in an unsupervised fashion. As a result, the generated offers originate from a space that is semantically aligned with the customer persona.

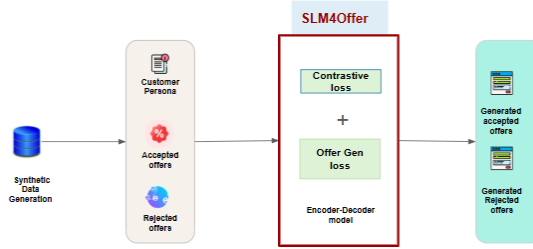

**Fig. 2.** Training workflow of SLM4Offer using contrastive learning with InfoNCE loss

Figure 2 provides a high-level overview of the fine-tuning process, where the model takes a customer persona as input along with previously accepted and rejected offers. The encoder processes the customer persona to generate a customer embedding. This embedding serves as the contextual input for the decoder, which generates candidate offers that the customer is likely to accept or reject. In addition to traditional supervised learning using cross-entropy loss, contrastive learning with InfoNCE loss is applied in the embedding space to enable unsupervised training.

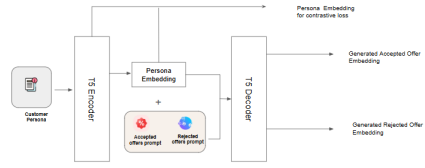

**Fig. 3.** Embedding generation from customer persona using the encoder of SLM4Offer

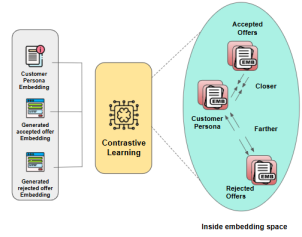

**Fig. 4.** Contrastive learning workflow of SLM4Offer using InfoNCE loss

Figure 3 illustrates the process of generating embeddings. The customer persona is passed through the encoder to produce a persona embedding. This embedding, along with the offer generation prompt, is fed into the decoder to extract embeddings of accepted and rejected offers from the decoder's hidden states. Figure 4 shows the contrastive learning-based fine-tuning process, which uses the persona embedding, the generated accepted offer embedding, and the generated rejected offer embedding. These embeddings are used to fine-tune the T5 model weights through contrastive learning. Figure 5 illustrates the use of traditional cross-entropy loss by comparing generated offers with ground truth offers. This loss is used for supervised fine-tuning of the model, ensuring that the generated offers are well aligned with the actual offers.

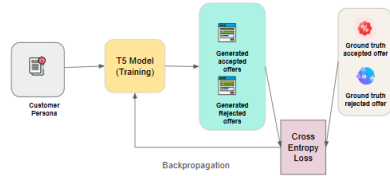

**Fig. 5.** Supervised fine-tuning of SLM4Offer using cross-entropy loss

## 5  Experimental Setup

This section describes the methodology used for synthetic data generation to fine-tune the T5 model for building SLM4Offer, as well as the pipeline used for model evaluation. The performance of SLM4Offer is compared with a baseline approach in which the T5 model is trained using Supervised Fine Tuning (SFT). The model's outputs are assessed using a large language model (LLM) as a judge. In addition, a human evaluation is conducted on a sample of the results. Hypothesis testing is then used to verify the reliability of the LLM-based evaluations.

### 5.1  Synthetic Data Generation

SLM4Offer is fine-tuned and evaluated using synthetic data. To mitigate biases during synthetic data generation, the generated customer personas and offers are based on statistical patterns observed in real company data. Diversity across age, interests, and behavioral types was ensured to simulate a balanced population. Customer personas and corresponding offers have been generated using Azure OpenAI LLMs. Each persona is defined by a combination of financial and behavioral attributes, including age (18–70), gender, monthly income ($30k–$200k), interests (e.g., Finance, Travel), spending patterns (e.g., Budget-conscious, High-spender), preferred payment methods, and financial goals. These attributes were sampled to create multiple unique personas. This customer persona data was fed to the LLM along with few-shot examples to generate offers. 3 accepted and 3 rejected offers were generated, which align with the persona's preferences.

### 5.2  Evaluation

The results of the proposed fine-tuning approach are compared against Supervised Fine-Tuning (SFT) performed on the same dataset. Offer acceptance rate which is defined as the ratio of accepted offers to the total number of generated offers,is used as the primary evaluation metric.

To assess the quality of the generated offers, an evaluation pipeline has been developed in which a large language model (such as GPT-4[1]) acts as a judge. Given: a) a customer persona, b) a list of previously accepted and rejected offers, and c) a newly generated offer, the LLM is prompted to classify whether the new offer is more likely to be accepted or rejected.

To validate the reliability of the LLM-based evaluation, a human assessment was conducted on a randomly selected subset of the test dataset. A group of annotators independently labeled each generated offer as accepted or rejected based on the corresponding persona and past offer history. An offer was then classified based on the majority vote of these annotations. These human labels were statistically compared with the LLM's classifications to measure agreement and establish a performance baseline for automated evaluation. To measure the association between the two, the Chi-Square Test of Independence was applied.

---

[1] Using Azure OpenAI Service

## 6   Results and Analysis

The generated synthetic dataset consists of 24,000 samples. For the experiments, the data has been divided into train, validation and test sets consisting of 21600, 2400 and 1000 samples respectively. The train set has been utilized for the finetuning of the model, with the validation set being used for monitoring the training, whereas the benchmarking on the experiments has been done on the test set. Following two fine-tuning methods are conducted on this dataset using the base model (Code-T5-Small):

1. Supervised Finetuning of the model with cross entropy loss
2. Finetuning the model with cross entropy loss and contrastive loss (InfoNCE loss)

Directly using the untrained base model gives the exact customer persona as the generated offer. This is because the pretrained model was trained on a large dataset and hence the model is unaware of the domain of the target dataset. To adapt the model to the domain of the dataset, finetuning of the dataset is required.

Figure 6 shows the loss plot for fine-tuning SLM4Offer. The loss initially fluctuates and stabilizes after a certain point, so the model from the 30th epoch has been used as the final trained model

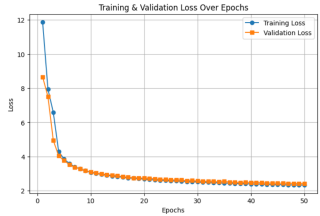

**Fig. 6.** Loss Plot for the training of SLM4Offer

**WeightWatcher** is a Python package for analyzing trained models and identifying training difficulties without accessing data. By fitting a power law to the ESD tail, it computes the exponent $\alpha$ to assess model quality. This helps detect issues in compression, fine-tuning, and generalization. Table 1 is the layer wise performance comparison between the two models. Alpha values less than 2 indicate overfitting whereas alpha values more than 6 indicate underfitting. The above observations indicate that SFT has more layers slightly in the underfitting zone compared to the contrastive model.

Table 2 shows the performance comparison between SFT and contrastive finetuning on the test set (1000 samples). There is a 17.5% increase in the acceptance percentage by using contrastive learning. This shows the superiority of contrastive learning in the use case.

**Table 1.** Layer-Wise Performance Comparison: SFT vs. Contrastive Model

|  | Ovefit | Normal | Underfit |
|---|---|---|---|
| **SFT** | 1 | 82 | 14 |
| **Contrastive** | 1 | 85 | 11 |

**Table 2.** Performance comparison between SFT and contrastive finetuning

|  | Offer accepted count | Offer Acceptance Rate (%) |
|---|---|---|
| **SFT** | 800 | 80% |
| **Contrastive finetuning** | 940 | 94% |

As mentioned in Section 5, whether the generated offer from the model would be accepted or rejected by the customer- has been evaluated using LLM as a judge using various prompting techniques including few shot prompting. Usage of LLM as a judge was verified with a set of human annotators on a sample of 200 records. Table 3 results of the evaluations by the two different type of judges:

**Table 3.** Evaluation Scores from Human Annotators and LLM Judges. 1 is for accepted and 0 is for rejected

|  | Human Evaluation 0 | Human Evaluation 1 |
|---|---|---|
| **LLM Evaluation 0** | 41 | 9 |
| **LLM Evaluation 1** | 3 | 147 |

The results of the chi-square test shows a statistically significant association between human and LLM evaluations, with a p-value less than 0.001. This indicates that the LLM can be taken as a judge to evaluate the generated offers and the results are reliable enough.

To summarize, the model was trained following the process illustrated in Figure 6, with a comparison of training dynamics between SFT and contrastive learning. The final model demonstrates improved performance, with a 17.5% increase in offer acceptance rate over the SFT baseline, as reported in Table 2. This improvement is also reflected in Figure 1, which highlights the model's ability to generalize and personalize offers effectively. These results suggest that incorporating contrastive learning alongside traditional fine-tuning leads to the generation of more robust, diverse, and persona-aligned marketing offers.

## Conclusion

This paper presents a novel contrastive learning based encoder-decoder framework for generating personalized marketing offers. By leveraging a fine-tuned T5 model and tweaking the embedding space between the encoder and the decoder using the InfoNCE loss, the paper's approach effectively aligns accepted offers with customer personas while distancing rejected ones. The system can periodically finetune the model using the most recent interaction data, along with the updated customer persona data, enabling it to adapt to user preferences dynamically. All the experiments have been done with respect to the financial domain, it can be extended to generate offers for any other domain such as e-commerce, retail, healthcare..etc. Future work may explore real-world deployment and expanding to adaptive personalization that evolves over time with user behavior. Currently, the paper uses customer personas as input to our contrastive learning-based framework. However, moving forward, each offer can be explicitly tagged with a corresponding product and campaign. This will enable better alignment with business goals. Such an integration will ensure that the generated offers are not only personalized, but also campaign-aware and product-specific.

## References

1. Arora, L., Singh, P., Bhatt, V., Sharma, B.: Understanding and managing customer engagement through social customer relationship management. J. Decis. Syst. **30**(2-3), 215–234 (Jul 2021)
2. Huang, H., Kajiwara, T., Arase, Y.: Definition modelling for appropriate specificity. In: Proceedings of the 2021 Conference on Empirical Methods in Natural Language Processing, pp. 2499–2509 (2021)
3. Kong, C., Chen, Y., Zhang, H.: Multitasking framework for unsupervised simple definition generation
4. Ni, K., Wang, W.Y.: Learning to explain non-standard english words and phrases. In: Proceedings of the Eighth International Joint Conference on Natural Language Processing, vol. 2, pp. 413–417. Short Papers, Taipei, Taiwan (2017)
5. Noraset, T., Liang, C., Birnbaum, L., Downey, D.: Definition modeling: Learning to define word embeddings in natural language. In: Thirty-First AAAI Conference on Artificial Intelligence (2017)
6. van den Oord, A., Li, Y., Vinyals, O.: Representation learning with contrastive predictive coding (Jul 2018)
7. Tomczyk, A.T., Buhalis, D., Fan, D.X.F., Williams, N.L.: Price-personalization: Customer typology based on hospitality business. J. Bus. Res. **147**, 462–476 (Aug 2022)
8. Zhang, H., Li, D., Yang, S., Li, Y.: Fine-grained contrastive learning for definition generation. In: Proceedings of the 2nd Conference of the Asia-Pacific Chapter of the Association for Computational Linguistics and the 12th International Joint Conference on Natural Language Processing (Volume 1: Long Papers). pp. 1001–1012. Association for Computational Linguistics, Stroudsburg, PA, USA (2022)